# AI systems and the reproduction of (standard) language ideologies in World Englishes

**Kingsley Ugwuanyi**
University of Nigeria, Nsukka; King's College London; SOAS, University of London

## Abstract

The rapid growth of large language models (LLMs) has resurrected age-old questions in sociolinguistics and world Englishes, such as who decides what counts as legitimate English, whose English is suspect etc. This paper examines how AI systems, their uses and discourse on them reflect, reinforce, and occasionally challenge (standard) language ideologies, which privilege Inner Circle norms and marginalize non-dominant Englishes. Drawing on evidence from empirical studies, media commentary, social media debates, and examples from AI outputs, the paper shows that AI technologies reproduce dominant language ideologies at different levels: training data, design protocols, evaluation benchmarks, user feedback and public commentary. The analysis uses the public controversy over AI-sounding language, especially the fixation on the word *delve*, to illustrate how speakers of English from the Global North police the English language norms of Global South English users. The paper also identifies what Christian Mair has called a "standardisation paradox": AI may homogenize English by privileging standard forms and at the same time pluralize Englishes through exposure to wide-ranging corpora and annotation work carried out by Global South users. In doing so, the paper argues that generative AI is reigniting long-standing debates in World Englishes about standardization, legitimacy, and the ownership of English, now playing out in algorithmic systems, model training, evaluation practices, and public discourse, where non-dominant Englishes are increasingly conflated with AI-generated speech. Discussing AI systems as a site where language ideologies are (re)produced, the paper argues for more inclusive design approaches that recognize the plurality of Englishes in order to address the real-world negative consequences of treating some as more legitimate than others.

**Keywords**: (standard) language ideologies; AI in linguistics research; World Englishes; linguistic hierarchies; machine-learning technologies

## 1. Introduction

The emergence of large language models (LLMs) has given fresh impetus to age-old sociolinguistic questions, such as "why is language standardized?", "who gets to decide what counts as correct or legitimate language?", and so on. The increasing development and deployment of AI tools requires us to interrogate the implicit ideas about language which are encoded in their design and outputs. These developments also speak to current debates about whether machine-learning technologies are reshaping how linguistics conceptualizes language or how inherited assumptions are being reproduced. This paper examines how (standard) language ideologies are reproduced or challenged in the design and use of LLMs, and what this means for how Englishes are researched and theorized in linguistics.

Language ideology refers to "sets of beliefs about language articulated by users as a rationalization or justification of perceived language structure and use" (Silverstein 1979: 193). In most cases, these beliefs determine how particular groups are perceived or valued, which makes it necessary for (critical) sociolinguists to be concerned about language-based ideologies in order to challenge their oft-prejudiced assumptions. Perhaps the commonest example of language ideology is what Lippi-Green (2012: 61) calls the "myth of the standard language", which is the belief that one variety of a language is inherently superior, more correct, or more legitimate than others (Erdocia et al. 2026; Milroy and Milroy 2012). These language myths are traditionally maintained through institutions (e.g., education) and cultural agents (e.g., media). However, emerging developments in digital technologies show that AI systems have become a space for enacting and maintaining language ideologies. In light of this, this paper argues that LLMs are not neutral linguistic tools but ideological actors. In different ways, AI systems reflect, reproduce, and sometimes even present an opportunity to challenge the dominant language ideologies underpinning English variation worldwide. Several studies have found that these systems tend to privilege standardized, Inner Circle Englishes as normative benchmarks, while treating other varieties as deviant, unintelligible, or less trustworthy (Dai and Hua 2025; Erdocia et al. 2026; Fleisig et al. 2024; Jones 2025; Rodriguez Louro 2025; Ugwuanyi et al. 2026).

This shift has important implications for World Englishes research. Traditionally, World Englishes scholarship has focused on documenting structural and pragmatic variation, but has been slower to examine how global technological infrastructures contribute to perceptions of legitimacy, correctness, and ownership of English(es). The rise of LLMs, therefore, introduces new perceptions and understandings of language, in which automated systems increasingly mediate what counts as acceptable English, thereby potentially changing both language practices and the objects of World Englishes research.

Considering generative AI as a site for the (re)production of language ideologies, this paper contributes to a growing body of critical sociolinguistic work that interrogates how AI systems encode, enforce, or subvert sociolinguistic hierarchies, ideologies, and attitudes. The analysis draws on a diverse range of material, including empirical studies of LLM design and outputs, as well as online comments, news articles, and social media discourse on AI-generated language. By examining both AI outputs and metalinguistic commentary surrounding them, the paper traces how beliefs about English(es), especially what counts as "legitimate", are reinforced or challenged across public domains. In line with the aims of this special issue, this paper asks whether AI-driven language technologies are reconfiguring how English language variation is conceptualized and analysed within linguistic research.

## 2. AI and the (re)production of standard language ideologies

LLMs are trained on immense and often undisclosed datasets, including web text, news media, academic content, and social media interactions (Kaubrė 2024). While these wide-ranging, mostly open-access sources might suggest inclusivity, there is an avalanche of evidence that AI outputs tend to privilege dominant, standardized forms of English from the Inner Circle (i.e., varieties of English spoken in majority English-speaking countries such as the UK and the US), thereby continuing to perpetuate the myth of standard language (Lippi-Green 2012). This myth frames the so-called standard variety as neutral, universally intelligible, and appropriate across all contexts, which, by the same token, renders non-standard or non-dominant varieties illegitimate and marked in many negative ways. As Milroy and Milroy (2012) argue, standardization is not a natural linguistic evolution but a politically, economically and ideologically driven process. In the case of AI, this process is now embedded in algorithmic logic.

This ideological bias begins with the training data. Large-scale training datasets, often drawn from sources like Common Crawl or Wikipedia, overrepresent formal and Inner Circle varieties of English. As Erdocia et al. (2024) observe, this reproduces a "dataist" view of language: language becomes measurable, extractable, and usable only when it conforms to certain, standardized formats. Evaluation practices in natural language processing further entrench these hierarchies. For example, it has been found that widely used benchmarks such as GLUE and SuperGLUE target a very specific subset of American English (Raji et al. 2021), producing inherently subjective outcomes despite their promise of objectivity. This reflects the situation in other domains of AI where datasets are overwhelmingly sourced from the Global North, including those that promise to explicitly include other varieties or languages. For instance, Shankar et al. (2017) show that although ImageNet set out to be inclusive by translating its image-search queries into multiple languages (Chinese, Spanish, Dutch, English, and Italian), the resulting dataset still drew over 45% of its images from the United States and more than 60% from a handful of Western countries. By contrast, China and India, which together account for over a third of the world's population, contributed barely 3% of the images. The authors argue that this lack of geographic, linguistic and cultural diversity led to poorer model performance on images from underrepresented cultures and languages. This failed attempt at inclusivity underscores how AI projects that claim to embrace diversity end up reproducing Global North dominance, including linguistically.

Such Global North or standard language ideologies are not only encoded through training data and evaluation but are also reinforced through the socio-technical choices in model design. The fine-tuning stage, including reinforcement learning with human feedback (RLHF), is typically carried out by annotators instructed to prioritize correctness, accuracy, and consistency, which are used to promote standard English norms. This aligns with Erfani's (2026) concept of "subjectification", through which ideology does not impose but "formats". RLHF processes, he argues, make annotators and users internalize the system's assumptions about what "good language" is, even when those assumptions marginalize their own linguistic repertoires. In doing so, they become participants in reproducing the very linguistic hierarchies and ideologies that characterize their language or variety as inferior.

In this sense, AI systems do not simply reproduce linguistic hierarchies; they recruit subjects into maintaining them. Ideology thus enters the dataset at the moment of selection, filtering, quality control, and alignment. As Erfani (2026) further explains, this normalization is analogous to standard language ideology itself, which naturalizes specific linguistic norms by ensuring they are embedded in everyday institutional and communicative practices. What emerges, then, is a technological reproduction of the ideology of the standard language. The resulting system privileges linguistic forms historically linked to educated, white, Global North speakers, even when exposed to more diverse forms of English. As a result, the authority of the standard is both hidden and reinforced through the practices and ideologies of those who train, test, and use AI tools.

The consequences of these ideologies become even more evident in the actual linguistic behaviour of LLMs. Recent evaluation studies have highlighted measurable differences in how LLMs process different varieties of English. In Fleisig et al. (2024), for example, GPT-4 and GPT-3.5 were tested across 10 varieties of English. The findings revealed a consistent pattern: responses to non-standard varieties triggered 19% more stereotyping, 25% more demeaning content, and 9% more comprehension failures compared to responses to Standard American English, which reveals how the ideology of the standard latent in training becomes active in performance.

This is further illustrated in Lin et al.'s (2024) evaluation of the reasoning of four LLMs, namely GPT-4o/4/3.5-turbo, LLaMA-3.1/3, Mistral, and Phi-3. The study evaluated LLM reasoning using prompts in African American Vernacular English (AAVE) and found that "compared to Standardized [American] English, almost all of these widely used models show significant brittleness and unfairness to queries in AAVE" (Lin et al. 2024: 1). Even when prompts were semantically and structurally equivalent to Standard American English, the LLMs struggled to reason appropriately in AAVE, often providing irrelevant, evasive, or incoherent responses. These findings show that these models are not merely unfamiliar with non-standard input but also that they advance the ideological stance that treats non-standardized varieties of English as deviant or inferior.

Even when LLMs expressly claim to give output in non-standard Englishes, the results reveal manifestations of standard language ideology. In a prompting experiment reported by Maria Kuteeva (Ugwuanyi et al. 2026), ChatGPT-4 was asked to reproduce an academic abstract in Nigerian English. The resulting text exaggerated lexical and grammatical features associated with Nigerian English by heavily drawing on elements of Nigerian Pidgin. Christian Mair's reflection in the same dialogue highlights that this problem extends even to more established national standards. According to him, ChatGPT often reverts to the norms of Standard American English, despite being asked to keep its output in Standard British English. These examples show that current LLMs not only essentialize non-dominant Englishes but also marginalize even well-codified varieties, thereby reasserting the dominance of a single, idealized standard English on a global level.

The examples discussed here represent only a fraction of the growing body of studies demonstrating how the design and use of LLMs contribute to the normalization of standard language ideology. These show that linguistic exclusion in AI systems is not incidental but systemic, which starts with the training data and manifests in AI outputs. However, these tendencies are not limited to technical design; they are also reflected in how AI-generated language is perceived, judged, and policed in everyday interaction. The next section turns to this broader concern by showing how standard language ideologies are reproduced and contested in public discourse about AI language.

## 3. Public discourse, language policing, and normative resistance

As stated above, beyond LLM design and outputs, (standard) language ideologies are also sustained and negotiated in the public discourse, where reactions to AI-generated language reveal broader attitudes towards languages or language varieties and their speakers. These reactions often perform a gatekeeping function by (re)echoing long-standing ideologies of standard English under the guise of "correctness". In this sense, public discourse on AI systems becomes another site where the myth of the standard language is reproduced and sometimes contested, as some of the cases presented below demonstrate.

A classic illustration of this process is the ongoing debate around the word *delve*, which has become a convenient marker of "AI-sounding" English (see Ugwuanyi et al. 2026). Several online commentators and news sources have claimed that a number of words, such as *delve*, *explore*, and *tapestry*, are being overused by ChatGPT. For example, based on a dataset of 50,000 ChatGPT responses, AI Phrase Finder (2024) listed the top 10 most common words used by ChatGPT, respectively: *explore*, *captivate*, *tapestry*, *leverage*, *embrace*, *resonate*, *dynamic*, *testament*, *delve*, and *elevate*. Despite *delve* ranking ninth in this list and even lower in other similar rankings (e.g., it ranked thirty-sixth in GPTZero's 2024 ranking; Constantino 2024), the word has been singled out by many commentators for criticism. And the reason for this is ideological.

A brief corpus check using the NOW corpus (2010–present; Davies 2013) suggests that the construction *delve into* was relatively stable from 2018 to 2022, fluctuating between 1.47 and 1.69 occurrences per million words, before rising sharply in 2023 (4.68 pmw) and remaining elevated in 2024 (4.38 pmw). This near tripling of normalized frequency coincides temporally with the widespread uptake of generative AI writing tools, and is consistent with the possibility that such systems amplify already available written-register patterns rather than introducing entirely new forms. At the same time, distribution across national components in NOW indicates that the highest frequencies occur in several Asian Englishes (e.g., Malaysia, Hong Kong, India), while Nigerian and Ghanaian components do not occupy the top positions. The salience of *delve* in public discourse is therefore better understood as an instance of indexical projection than as evidence of national ownership of a distinctive lexical feature. These usage patterns make the gap between empirical distribution and public attribution more visible.

Despite the above evidence from corpus data, Hern (2024), in an article published in *The Guardian*, described the use of *delve* by ChatGPT thus:

> I said “delve” was overused by ChatGPT compared to the internet at large. But there’s one part of the internet where “delve” is a much more common word: the African web. In Nigeria, “delve” is much more frequently used in business English than it is in England or the US. So the workers training their systems provided examples of input and output that used the same language, eventually ending up with an AI system that writes slightly like an African.
>
> And that’s *the final indignity* [my emphasis]. If AI-ese sounds like African English, then African English sounds like AI-ese.

Hern’s conclusion rests on a claim that is not supported by corpus evidence and reproduces a centre–periphery hierarchy that is prejudiced against African Englishes. To call it “the final indignity” shows that he thinks it is undesirable for AI to “sound African”. The phrase reproduces the very hierarchy it seeks to criticize: it implies that African English is legitimate only when it does not resemble machine language. This moment in public discourse shows how dominant language ideologies can re-emerge in critical commentaries on AI language.

A similar stance is evident in a widely circulated LinkedIn post by Rohit Aggarwal (2024), a professor and AI researcher at the University of Utah, entitled *Let us "delve" into the influence of Nigerian English on ChatGPT and Global Writing Patterns.* Aggarwal writes:

> Let us "delve" into the influence of Nigerian English on ChatGPT and Global Writing Patterns
>
> With a large pool of Nigerian annotators providing key inputs to refine ChatGPT's natural language abilities, it's intriguing to consider how these linguistic quirks could be seeping into the AI's outputs and influencing the writing of its millions of users worldwide. Will we start to see more "globalisms" and non-standard usages entering the mainstream English lexicon as a result of ChatGPT's growing popularity? How might this impact formal vs. casual writing and communication norms over time?
>
> I wonder if ChatGPT's overly polite language for emails may be indeed influenced by its Nigerian annotators. Elaborate greetings, praise, and expressions of goodwill may be common in Nigerian culture.
>
> What do you think about ChatGPT's generated content? Have you toned down the flowery outputs or removed peculiar words frequently used by ChatGPT? Let me know your experiences in the comments!
>
> So next time you see high usage of words like "delve" you know where it came from.

First, the title of the post is ideologically loaded. It does not simply identify a lexical feature; it presents that feature as evidence of Nigerian English influence on ChatGPT and, by extension, on global written usage. This presupposes a causal relationship between a national variety of English and the spread of a particular lexical item in AI-generated writing. Although Aggarwal appears to present this as a neutral observation, the post reproduces the same assumptions evident in Hern's article: African Englishes are treated as "linguistic quirks" and a deviation from an assumed mainstream standard. The warning that users may have to "tone down flowery outputs or remove peculiar words" produced by ChatGPT, influenced by its Nigerian annotators, depicts Nigerian English, and by extension other Global South Englishes, as an undesirable presence contaminating English. His closing remark, "next time you see high usage of words like 'delve,' you know where it came from", reinscribes a boundary between the mainstream or standard varieties of English and its supposed peripheries. Like Hern, Aggarwal gestures towards inclusivity while ultimately reaffirming a centre–periphery hierarchy in World Englishes. This reveals how even public discussions about AI's behaviour are saturated with language ideologies that devalue English norms of the Global South, as what counts as a "quirk" or "peculiarity" depends on who gets to define the norm.

This attitude towards *delve* was intensified by Paul Graham (2024), a prominent American investor and essayist, who tweeted:

> Someone sent me a cold email proposing a novel project. Then I noticed it used the word "delve."
> My point here is not that I dislike "delve," though I do, but that it's a sign that text was written by ChatGPT.

The tweet, which quickly went viral (with over 17 million views), presented *delve* as a global symbol of "inauthentic" English and as proof of deception by suggesting that anyone who writes that way must be a machine or, by implication, a non-native speaker.

The backlash from Global South English users, as reported by *The Cable* (Bodunde 2024), revealed how such policing reinscribes linguistic inequality. The report highlighted several responses that challenged Graham's assumptions about the contested lexical item and linked his claim to wider histories of language policing, colonial education and suspicion towards non-Inner Circle English users. Some of the responses, all taken from Bodunde (2024), are shown below:

- @prateeks: A lot of folks who grew up in countries where english was not spoken actually use a lot of 'written-first' words in their writing because that is how they learned the language.
- @AdaErema: Imagine assuming a paper is written by AI just for using certain English words. Most of us were taught to use these words in school. The fact some words aren't common in American English doesn't mean it's the same everywhere. Nigerians used most of these “AI-generated words” in formal writing before the invention of AI. Open up your mind to other cultures. This shows the bias International students face.

- @elnathan_john: What Paul Graham did is dangerous and colonial. Many of us were flogged for using vernacular in school. We are treated with suspicion when we travel to the west, made to prove our knowledge many times over. Our knowledge is doubted and belittled.
- @ozzyetomi: Paul Graham’s tweets today highlight the implicit bias foreign students face a lot with white teachers and professors. They already think you’re cheating half the time just cos you’re smart, now you’ll say ‘delve’ and ‘unlock’ and they will say you’re using chatGPT to write.
- @UjuAnya: Paul Graham’s absurd claim that using words like “delve” indicate ChatGPT reveals his assumption that everyone shares his linguistic limitations. Like US and UK professors who punish Africans for using English their limited knowledge tells them such students can’t possibly know.

These responses reframe the controversy as a dispute over linguistic authority: who is entitled to judge English, whose usage is treated as ordinary, and whose lexical choices are recast as artificial, suspicious or deficient. @prateeks’s comment points to the “written-first” acquisition route through which many users outside the Inner Circle encounter English, while @AdaErema explicitly challenges the assumption that American English provides the measure of English usage worldwide. @elnathan_john and @ozzyetomi extend the issue to colonial schooling and institutional suspicion, showing how speakers from formerly colonised contexts are often made to prove the legitimacy of their knowledge and language. @UjuAnya’s response makes the same point in relation to academic gatekeeping, where African students’ command of English may be questioned not because their usage is deficient, but because it exceeds the linguistic expectations of those positioned to evaluate them.

Such responses re-appropriate *delve* as part of legitimate usage well established in Nigerian English and other Global South Englishes before the advent of GenAI. More importantly, they show how AI technologies are reopening age-old disputes over authority in English: who gets to decide what counts as legitimate usage, whose English is treated as suspect, and whose linguistic competence is recognised. We know that what is considered as “correct”, “fluent” or “authentic” is not merely a linguistic matter, but is determined by speaker position, institutional authority and geopolitical power (Lippi-Green 2012; Milroy & Milroy 2012). As Woolard (1998) notes, language ideologies are inseparable from speaker positions; here, Graham and those who share his assumptions act as evaluators of linguistic legitimacy, while Global South speakers resist that evaluation.

This “delve controversy” therefore illustrates how public discourse mirrors ideologies already embedded in AI design. A single lexical item becomes the object around which a wider debate about who owns English, who sounds legitimate, and who decides what counts as standard is articulated. The policing of *delve* by speakers of English from the Inner Circle was never linguistic; it was ideological. At the same time, the collective pushback from speakers of Global South Englishes shows that such hierarchisation can be questioned in the same digital spaces that sustain it. This leads to the next point: that generative AI, while helping to sustain the myth of the standard language, may also create openings for challenging linguistic inequalities.

## 4. The standardization paradox and the potential for inclusivity

Despite the exclusionary tendencies of AI systems discussed above, they also hold a lot of potential for linguistic inclusivity. This reflects what Mair (2023, 2025) refers to as the "standardisation paradox" in the AI era, the fact that English is becoming both more homogeneous and more heterogeneous at the same time. On the one hand, dominant ideologies of standardness continue to shape model design, benchmark evaluations, and user expectations. On the other hand, the very scale of LLMs, which use massive, uncurated datasets from wide-ranging sources, exposes them to a wider range of Englishes from across the world.

Part of this exposure arises from the inclusion of user-generated content such as Reddit posts, Stack Exchange discussions, social media posts, and YouTube transcripts, where diverse Englishes are present. Thus, linguistic variation "seeps into" (to use Aggarwal's phrase) training data. However, this "seeping into" does not necessarily mean legitimization, as the motivation for employing annotators from the Global South is largely economic: they are paid significantly less than their Global North counterparts (Hern 2024). For example, a study involving AI data annotators from India, Kenya, Nepal, the Philippines, Nigeria, and the US reveals that annotators feel constrained by strict quality control instructions based on Western linguistic and cultural norms, which tend to clean out their own English repertoires (Wang et al. 2022). Yet, despite these constraints, these annotators' linguistic repertoires inevitably influence the data they handle. Even if this is not yet significant.

This raises a key question about what kind of inclusion AI systems can realistically achieve. Even if speakers of Global South Englishes shape aspects of model structure, the "voice" that users ultimately hear in audio-based AI tools remains standardized and Western. Michel et al. (2025) observe that users from the Global South often feel alienated by the default voice in most AI systems, which they describe as "not a representation of me".

However, within these contradictions lie possibilities for inclusion. Some recent efforts in natural language processing research underline this possibility. Held et al.'s (2023) Multi-VALUE project, for instance, uses small, swappable components called adapters that can fine-tune existing models for particular dialects or regional varieties without retraining them from scratch. This approach opens up the possibility of more equitable model performance across Englishes, as well as shows that inclusivity need not come at the cost of technological efficiency.

A more fundamental shift, however, requires a change in orientation, beliefs, and practices. AI developers, researchers, and corporations must first acknowledge that English is not a single language, but a combination of legitimate varieties shaped by social, historical, and political forces. This recognition must be followed by deliberate inclusion in model design and data curation, for example, by building parallel datasets that represent multiple Englishes and moving beyond Inner Circle ideologies that continue to define the linguistic mainstream. Another important step forward would be to review data-cleaning and data-generation processes because they contribute to erasing variation.

## 5. Conclusion

This paper has shown that LLMs are not neutral tools, but ideological actors that reproduce entrenched hierarchies within English. The evidence from AI design and outputs and public discourse demonstrates that non-dominant varieties of English spoken chiefly in the Global South are often treated as deviations from a presumed standard. The discrepancy between corpus-based distribution and public attribution in the case of *delve* further shows how language ideologies are often not based on empirical usage, but instead emerge from what people think of certain populations of language users. While Mair's (2023, 2025) "standardisation paradox" suggests that English in the age of AI has the tendency to become both more homogeneous and more diverse, it remains unclear how far this inclusivity potential can go, given that current AI systems are heavily underpinned by language norms of the Global North.

These language ideologies, which underlie much of the AI ecosystem, deserve closer scrutiny from critical sociolinguists because "dialect prejudice has the potential for harmful consequences: language models are more likely to suggest that speakers of AAE be assigned less-prestigious jobs, be convicted of crimes and be sentenced to death" (Hofmann et al. 2024: 147). Further, as Rodriguez Louro (2025) asks:

> If an AI tutor fails to understand a Nigerian English construction, who bears the cost? If a job application written in Indian English is marked down by an AI-powered resume scanner, what are the consequences? If an Australian First Nations elder's oral history is transcribed by voice recognition software and the system fails to capture culturally significant terms, what knowledge is lost or misrepresented?

Such questions remind us that linguistic hierarchization is not an abstract concept but a material condition that can determine who is understood, who is excluded, who is hired, who receives social services and whose knowledge is recorded or erased (Irvine and Gal 2000).

For World Englishes, then, generative AI is not merely another context in which English varies; it has become a new site in which long-standing debates about standardization, legitimacy, and ownership of Englishes are being reopened. If World Englishes scholarship has traditionally examined these questions in relation to speakers, texts, and institutions, it must now also confront them in relation to datasets, design principles, benchmarks, model outputs, and the public discourse through which hierarchies among Englishes are reproduced and contested. Addressing these inequities requires that linguists, technologists, and institutions recognize language ideologies as a matter that must be considered in the design of AI systems. The challenge ahead is to ensure that future AI systems do not simply reproduce the linguistic hierarchies of the past but contribute to a more inclusive and pluralistic understanding of Englishes in the world.